# Additive Belief-Network Models

**Paul Dagum**
Section on Medical Informatics
Stanford University School of Medicine
and
Rockwell Palo Alto Laboratory
444 High Street
Palo Alto, California 94301

**Adam Galper**
Section on Medical Informatics
Stanford University School of Medicine
Stanford, California 94305-5479

## Abstract

The inherent intractability of probabilistic inference has hindered the application of belief networks to large domains. *Noisy OR-gates* [30] and *probabilistic similarity networks* [18, 17] escape the complexity of inference by restricting model expressiveness. Recent work in the application of belief-network models to time-series analysis and forecasting [9, 10] has given rise to the *additive belief-network model (ABNM)*. We (1) discuss the nature and implications of the approximations made by an additive decomposition of a belief network, (2) show greater efficiency in the induction of additive models when available data are scarce, (3) generalize probabilistic inference algorithms to exploit the additive decomposition of ABNMs, (4) show greater efficiency of inference, and (5) compare results on inference with a simple additive belief network.

## 1  INTRODUCTION

A *Bayesian belief network* is a model that employs a graphical structure to characterize a joint-probability distribution. The research that culminated in the development of belief networks can be traced to the work of Lewis [28]. Lewis addressed the problem of approximating an arbitrary distribution with low-order distributions by measuring the "goodness" of the approximation with the Kullback-Liebler cross-entropy measure [26]. Subsequent research focused on methods for choosing a set of low-order distributions that best approximates the original distribution [4, 13, 25]. This research neglected the implicit dependencies and independencies that structure a domain, and instead sought to provide an approximation of the true distribution with minimal cross entropy. For domains in which structural dependencies are known, Pearl [29, 30] characterized belief networks as probabilistic models that separate *structure learning*, or the determination of model dependencies, from *parameter learning*, or the determination of the probabilities that quantify the dependencies.

The recent development of belief-network applications [1, 2, 19, 22, 24, 32] has stimulated the maturation of techniques for probabilistic inference in belief networks and for induction of belief networks [7, 11, 23, 20, 21, 27, 30, 31]. It is now evident that the intractability of available probabilistic inference algorithms hinders the application of belief networks to large domains. Both exact and approximate probabilistic inference is **NP**-hard, and therefore, we do not hope to find tractable solutions to inference in large belief networks [6, 12].

The formal proofs of the complexity of inference have spurred the development of approximate modeling techniques that restrict the form of the model in order to reduce the complexity of inference. Examples include *noisy OR-gates* [30], used in QMR-DT [32], and *probabilistic similarity networks* [18, 17] used in Pathfinder [19]. Motivated by recent developments in belief-network models for time-series analysis and forecasting [9, 10], we introduce a new approximate modeling methodology: the *additive belief-network model (ABNM)*. We (1) discuss the nature and implications of the approximations made by an additive decomposition of a belief network, (2) show greater efficiency in the induction of additive models when available data are scarce, (3) generalize probabilistic inference algorithms to exploit the additive decomposition of ABNMs (4) show greater efficiency of inference, and (5) compare results on inference with a simple additive belief network.

In [8], we develop a theory of *algebraic belief-network models* that extends the expressivity of ABNMs by allowing *multiplicative* decompositions of a belief network. For this extended class of models, we show results for learning and probabilistic inference similar to those presented here for ABNMs.

## 2  ADDITIVE MODEL

The theory of nonparametric additive models is relatively recent [3, 14, 15]. Before we define ABNMs, we discuss briefly *additive models* and *generalized additive*

*models.*

## 2.1 ADDITIVE MODELS AND GENERALIZED ADDITIVE MODELS

Suppose we desire to model the dependence of a variable $Y$ on variables $X_1, ..., X_p$. We wish to do so for purposes of (1) *description*, to model the dependence of the response on the predictors in order to learn more about the process that produces $Y$, (2) *inference*, to assess the relative contribution of each predictor to $Y$, and (3) *prediction*, to predict $Y$ given values of the predictors. When linear regression of $Y$ on $X_1, ..., X_p$ provides an adequate model, its simplicity makes it the preferred method. The inadequacy of linear regression, for example, in medical domains [16], led to the development of additive models.

Additive models maintain the attractive properties of linear-regression models; they are additive in the predictor effects, but are not constrained by assumptions of linearity in the predictor effects. An additive model is defined by

$$\mathbf{E}(Y|X_1, ..., X_p) = \sum_{i=1}^{p} f_i(X_i), \qquad (1)$$

where the functions $f_i$ are arbitrary.

Generalized additive models extend additive models in the same way that generalized linear models extend linear models: they allow a general *link* between the predictors and the dependent variable. For example, log-linear models for categorical data and gamma-regression models for responses with constant coefficient of variation represent generalized linear models. These generalizations extend naturally to additive models [14].

## 2.2 ADDITIVE BELIEF-NETWORK MODELS

We first define formally a belief-network model. A belief network consists of a directed acyclic graph (DAG) and a set of conditional probability functions. Let $X_1, ..., X_n$ represent the nodes of the DAG, and let $\pi(X_i)$ denote the set of parents of $X_i$ in the DAG. The nodes of the DAG represent the variables of the belief network. The directed arcs in the DAG represent explicit dependencies between the variables. We assume that each variable is binary valued. To complete the definition of a belief network, we specify for each variable $X_i$, an associated conditional probability function (table) denoted $\Pr[X_i|\pi(X_i)]$. The full joint probability distribution is given by [30]

$$\Pr[X_1, ..., X_n] = \prod_{i=1}^{n} \Pr[X_i|\pi(X_i)]. \qquad (2)$$

The key link between the theory of additive models presented in Section 2.1 and ABNMs lies in the interpretation of the conditional probability functions. For node $X_i$, specification of the function $\Pr[X_i|\pi(X_i)]$ implies a nonparametric model of the effect of the predictors $\pi(X_i)$ on the dependent variable $X_i$. To make clear the analogy with additive models, we let $Y$ denote $X_i$, and we assume that $\pi(X_i) = \{X_1, ..., X_p\}$. We might define an additive belief-network model to be a model that satisfies

$$\Pr[Y|X_1, ..., X_p] = \sum_{i=1}^{p} \alpha_i \Pr[Y|X_i], \qquad (3)$$

where the weights $\alpha_i$ sum to one, and thereby normalize the summation expression. Equation 1, which expresses the property of additive models, follows directly from Equation 3:

$$\mathbf{E}[Y|X_1, ..., X_p] = \sum_{i=1}^{p} f_i(X_i),$$

where

$$f_i(X_i) = \alpha_i \sum_{y} y \Pr[Y = y|X_i] = \alpha_i \mathbf{E}[Y|X_i].$$

The expectation $\mathbf{E}[Y|X_i]$ is with respect to the distribution $\Pr[Y|X_i]$.

We define ABNMs to be more general than additive models. We allow additive interaction terms, such as $\Pr[Y|X_i, X_j]$, in Equation 3. Thus, an ABNM has conditional probabilities that satisfy

$$\Pr[Y|X_1, ..., X_p] = \sum_{i=1}^{k} \alpha_i \Pr[Y|S_i], \qquad (4)$$

where $S_i$ are subsets of the predictors such that $S_1 \cup S_2 \cup \cdots \cup S_k = \{X_1, ..., X_p\}$. We note that in general, it is *not* necessary that the $S_i$ form a disjoint partition of the predictors $X_1, ..., X_p$. Thus, for example, $\Pr[Y|X_i, X_j]$ and $\Pr[Y|X_i, X_k]$ might be two valid interaction terms in the expression for $\Pr[Y|X_1, ..., X_p]$. In Section 3, we show that it is frequently necessary to allow for nondisjoint partitions of the predictors if we are to arrive at a coherent semantics for the additive decomposition.

Whereas specification of the univariate functions in additive models is accomplished easily through a recursive *backfitting algorithm* [16], specification of interaction terms is complicated by the numerical instability and biases of the fitting procedure in higher dimensions. Induction of the additive interaction terms in ABNMs is relatively free of the complications we encounter in backfitting interaction terms in additive models. We discuss induction of the additive terms in ABNMs in Section 4.

## 3 SIGNIFICANCE OF ADDITIVE DECOMPOSITION

The decomposition of $\Pr[Y|X_1, ..., X_p]$ into additive terms is an approximation of the true functional dependence of $Y$ on its predictors. The decomposition originated in the design of *dynamic network models (DNMs)*—belief-network models for time-series analysis and forecasting [9, 10]. If the decomposition is chosen with insight, then very little is lost in the approximation. Furthermore, there is much to be gained from an additive decomposition when the number of predictors $p$ for $Y$ is large—as is usual in large, complex domains. In Section 4, we discuss how decomposition facilitates the induction of belief networks from data, and in Section 5, we show that additive decomposition accelerates probabilistic inference. A carefully chosen additive decomposition of key conditional dependencies in a belief network— even if the decomposition has only two additive terms—can transform an intractable inference problem into a problem that is solved readily.

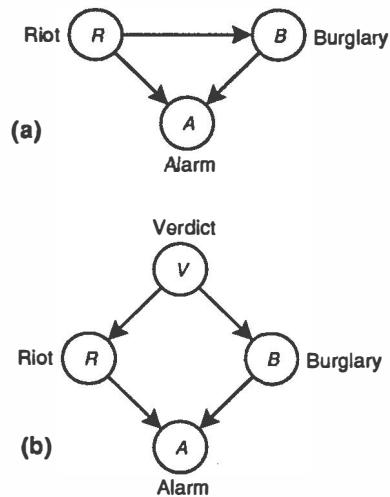

Figure 1: Two version of the Riot belief network. (a) An explicit intercausal dependence exists between Riot and Burglary. (b) An implicit intercausal dependence exists between Riot and Burglary, mediated through Verdict.

A poorly chosen decomposition, however, will introduce biases into the model through assumptions about independencies among predictors. For example, the decomposition

$$\Pr[Y|X_1, X_2] = \alpha \Pr[Y|X_1] + (1 - \alpha) \Pr[Y|X_2]$$

allows us to study the dependence of $Y$ on the predictor $X_1$, independent of the values of $X_2$, and to study the relative contribution of $X_1$ in explaining $Y$. This simplicity and insight come at the expense of assuming a specific type of interaction between $X_1$ and $X_2$, which at times, may be sufficiently significant to bias an inference adversely. Consider the following example modified from Pearl [30]. Let $Y$ represent the event of "triggering an alarm", $X_1$ represent a "burglary", and $X_2$ represent a "riot". We regard $X_1$ and $X_2$ as two sources of information regarding $Y$, the state of the alarm. Each source provides an assessment of the alarm through the terms $\Pr[Y|X_1]$ and $\Pr[Y|X_2]$. In an additive decomposition, we pool our sources of information through a weighted sum of the terms $\Pr[Y|X_1]$ and $\Pr[Y|X_2]$. Thus, our belief regarding the state of the alarm lies within the two extremes provided by each source of information acting independently. However, if both predictors $X_1$ and $X_2$ are independent, then learning about a riot *and* a burglary should increase our belief in the alarm beyond the belief predicted by either source alone. In general, the additive decomposition cannot model the synergy of two *independent* predictors $X_1$ and $X_2$ of $Y$. Elsewhere [8], we show how a multiplicative decomposition of conditional probabilities can model synergy between independent predictors. Furthermore, multiplicative decomposition, like additive decomposition, improves the efficiency of probabilistic inference and induction.

### 3.1 INTERCAUSAL DEPENDENCE AND ADDITIVE DECOMPOSITION

We have shown that additive decomposition should not be used when predictors are independent, but we haven't shown when an additive decomposition is appropriate. Suppose a riot and a burglary are not independent events. For example, there may be an explicit dependence, as depicted in Figure 1a, if we believe that burglaries are more likely to occur during a riot than during peaceful circumstances. A riot causes vandalism that might lead to burglary, thereby triggering the alarm, or vandalism might trigger the alarm directly. Alternatively, the dependence between a riot and a burglary might be implicit, as in Figure 1b, in which the probability of a riot or a burglary is high given the jury verdict in a high-profile trial. Borrowing from Wellman and Henrion [33], we refer to explicit or implicit dependence between causes of an event as *intercausal dependence*. Strong intercausal dependence between predictors—for example, a high probability of burglary given a riot—reduces the confidence gained by observing both a riot and a burglary. The two information sources share background knowledge, and therefore observing both causes does not necessarily increase our confidence in the state of the alarm. The weights of the additive decomposition represent our respective confidence in the two sources of information, based on, for example, the reliability of the observations, an assessment of predictive fidelity, or the consistency of our background knowledge. Thus, we can use the weights to adjust for dissonant information. If we learn about the riot through an established media channel, but our information regarding burglary is third hand, we can adjust the contribution of the burglary report to our belief in the state of the alarm by discounting the weight of its prediction.

## 3.2   A PRESCRIPTION FOR ADDITIVE DECOMPOSITION

In summary, when the predictors $\{X_1, ..., X_p\}$ of $Y$ exhibit pairwise intercausal dependence—that is, intercausal dependence for any pair of predictors $X_i$ and $X_j$—then the additive decomposition of Equation 3 is justified. More generally, we define the *intercausal dependence graph* for node $Y$ to consist of nodes $V = \{X_1, ..., X_p\}$ and undirected edges $E$ defined by the intercausal dependencies. When the intercausal dependence graph is a clique, then the additive decomposition of Equation 3 is justified. When the intercausal dependence graph is not a clique, then let $\{X_1, ..., X_k\}$ denote the vertices of the largest clique within the intercausal dependence graph. Let $N(X_i)$ denote the neighbor of node $X_i$ in the graph. The additive decomposition for $Y$ is now given by Equation 4, with $S_i = \{X_i\} \cup (V \setminus N(X_i))$, $i = 1, ..., k$. The set $(V \setminus N(X_i))$ denotes the predictors that are not intercausally dependent with $X_i$.

## 3.3   NEGATIVE PRODUCT SYNERGY AND ADDITIVE DECOMPOSITION

A property of the additive decomposition is that the *absolute* contribution of a riot to the probability of triggering the alarm is independent of whether or not the building is burglarized at the same time a riot occurs. A similar property holds for the absolute contribution of a burglary. In the formalism of Wellman and Henrion [33], it follows that for additive decomposition, the predictors exhibit *zero additive synergy* with respect to $Y$. However, the *relative* contribution of a riot to the probability of triggering the alarm *is* dependent on whether or not the building is burglarized at the same time. More generally, it is straightforward to show for the additive decomposition that if the predictors *positively influence* $Y$, then the predictors exhibit *negative product synergy* with respect to $Y$, and this result is independent of the choice of $\alpha$. In other words, under the additive decomposition, the proportional increase in the probability that the alarm rang due to learning of a riot is smaller when we know that the building was burglarized, than when we know that the building was not burglarized.

## 4   FITTING ADDITIVE BELIEF-NETWORK MODELS

In the additive decomposition expressed in Equation 3, we left unspecified the method of estimation of the weights $\alpha_i$. In this section, we discuss the significance of these weights and we present alternative methods for their estimation. Before we proceed, we compare the complexity of induction of the conditional probabilities $\Pr[Y|S_i]$ with the complexity of induction of $\Pr[Y|X_1, ..., X_p]$.

## 4.1   INDUCTION OF CONDITIONAL PROBABILITIES

When there are many predictor variables, we may overspecify the model with insufficient data if we attempt to specify $\Pr[Y|X_1, ..., X_p]$ directly. An overspecified model will produce biased inferences. For example, if $p = 10$ and each variable has four possible values, then we must specify $2^{20}$ probabilities for each value of the dependent variable $Y$. Not only is this beyond the realm of *any* domain expert, but it is clearly beyond the realm of belief-network induction algorithms [7, 31]. These algorithms induce the conditional probabilities by counting cases in a database of model instantiations. To guarantee reasonable convergence of the algorithm, for each instantiation of the predictors, we would like to observe at least ten cases in the database with this instantiation. Thus, we require a database of at least $10^7$ cases—clearly a prohibitive demand. On the other hand, even a single additive split of $\Pr[Y|X_1, ..., X_{10}]$ into two conditional probabilities, each with five predictors, reduces substantially the number of cases required for induction. With a single decomposition, we require specification of $2^{10}$, or one thousand, probabilities for each value of the dependent variable, and a database of $10^4$ cases will suffice for induction.

## 4.2   ESTIMATION OF PARAMETERS

In Equation 3, we argued that the weights $\alpha_i$ were necessary to normalize the sum of conditional probabilities. Although the $\alpha_i$ normalize the sum, they also affect significantly each summand's relative contribution to the conditional probability $\Pr[Y|X_1, ..., X_p]$. For example, in DNMs, we decompose conditional probabilities into two terms: the first term contains the subset of predictors that are contemporaneous with the dependent variable $Y$, and the second term contains the subset of predictors that are noncontemporaneous with $Y$. The weights $\alpha$ and $1 - \alpha$ in DNMs represent the relative contribution to the prediction of $Y$ from contemporaneous and noncontemporaneous information. A value of $\alpha$ near one favors the prediction of $Y$ based on contemporaneous data, whereas a value of $\alpha$ near zero favors the prediction of $Y$ based on noncontemporaneous data.

*Fitting* an ABNM refers to the specification of the weights. We view the weights as probabilities that denote the contribution of each summand to $\Pr[Y|X_1, ..., X_p]$. We fit ABNMs through iterative Bayesian update of the weights with new evidence.

Let $\Pr[\alpha_1, ..., \alpha_k] = \Pr[\vec{\alpha}]$ denote the probability distribution for the weights. Assume that we have observed evidence that consists of $m$ independent instantiations of the network $E_1, ..., E_m$, with the union of this evidence denoted by $\mathcal{E}_m$. Let $Pr[\vec{\alpha}|\mathcal{E}_m]$ denote the probability distribution for the weights $\alpha$ after we observe the evidence. We update the distribution with evi-

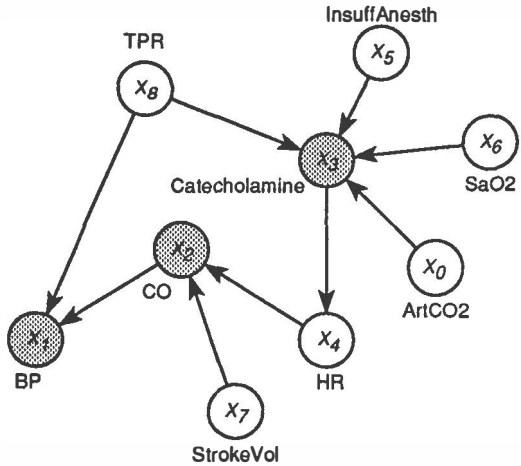

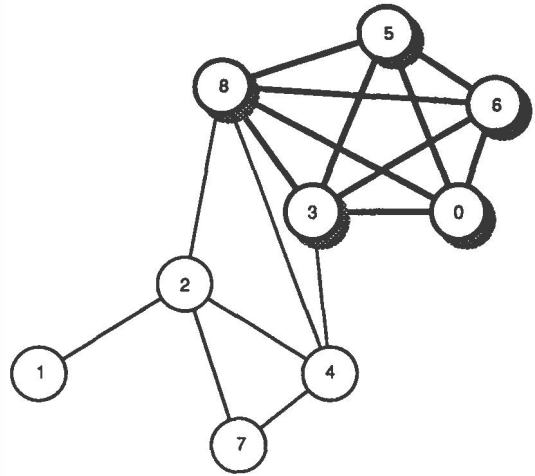

Figure 2: AlarmX, a subnetwork of the 37-node ALARM belief network designed for patient monitoring in an intensive care unit. Nodes $x_1$, $x_2$, and $x_3$ have additive decompositions. Node $x_1$ and $x_2$ have a unique partition of their parent nodes since they each have two parents. The parent set of $x_3$ is partitioned into two sets, $S_1 = \{x_0, x_6\}$ and $S_2 = \{x_5, x_8\}$. ArtCO2: arterial CO2 level, BP: blood pressure, CO: cardiac output, HR: heart rate, InsuffAnesth: insufficient anesthesia, SaO2: oxygen saturation, StrokeVol: stroke volume, TPR: total peripheral resistance.

Figure 3: Triangulation of AlarmX. The shadowed nodes form the largest clique.

dence $E_{m+1}$ according to Bayes' rule:

$$\Pr[\vec{\alpha}|E_{m+1}, \mathcal{E}_m] = {}^{\backprime}k\,\Pr[E_{m+1}|\vec{\alpha}, \mathcal{E}_m]\,\Pr[\vec{\alpha}|\mathcal{E}_m],$$

where $\Pr[E_{m+1}|\vec{\alpha}, \mathcal{E}_m]$ is the probability of evidence $E_{m+1}$ we compute with the ABNM and $k$ normalizes the distribution.

### 4.3 CROSS-ENTROPY VALIDATION

To *validate* an ABNM, we can measure how closely the inferences generated with the ABNM approximate the inferences generated by a full belief-network model. Let $\Pr[X_1, ..., X_n]$ and $\Pr'[X_1, ..., X_n]$ denote the full joint probability distributions of a belief network BN and of an ABNM that approximates BN. The Kullback-Liebler cross-entropy [26] measures how well the distribution $\Pr'[X_1, ..., X_n]$ approximates $\Pr[X_1, ..., X_n]$:

$$I_{\Pr,\Pr'} = \sum_{x_1,...,x_n} \Pr[x_1, ..., x_n] \log \frac{\Pr[x_1, ..., x_n]}{\Pr'[x_1, ..., x_n]}. \quad (5)$$

We can easily show that $I_{\Pr,\Pr'} \geq 0$, and it is equal to zero only if the distributions are identical. The closer $I_{\Pr,\Pr'}$ is to zero, the better $\Pr'[x_1, ..., x_n]$ approximates $\Pr[x_1, ..., x_n]$. Thus, once we specify the parameters of the ABNM, we can construct its full joint probability and compute the cross entropy. Unfortunately, to compute the cross-entropy, we must sum over all belief-network instantiations, and for large networks cross-entropy calculations are intractable.

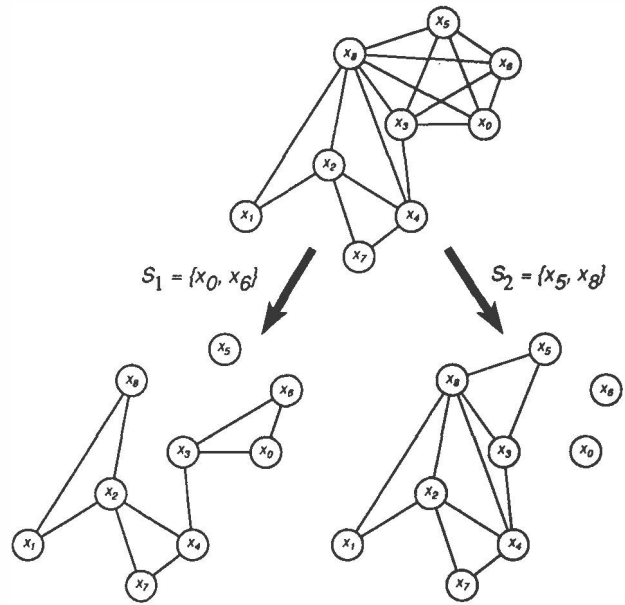

Figure 4: The dissection of AlarmX at node $x_3$. The dissection generates two new belief networks, which are shown in triangulated form. The triangulated network for partition $S_1$ is obtained by deleting arcs $(x_5, x_3)$ and $(x_8, x_3)$ from AlarmX and retriangulating. The triangulated network for partition $S_2$ is obtained similarly by deleting arcs $(x_0, x_3)$ and $(x_6, x_3)$. Dissection along node $x_3$ reduces the five-node clique in the triangulation of AlarmX into two three-node cliques.

The cross-entropy $I_{\text{Pr},\text{Pr}'}$ is a function of the parameters of the ABNM. Each node with an additive decomposition also has an associated set of parameters. If we assume that the parameters of a node can be specified independently of each other and of the parameters of other nodes, then we can compute the parameters that minimize $I_{\text{Pr},\text{Pr}'}$. For node $X_i$ with an additive decomposition, let $S_1^i, ..., S_k^i$ denote the additive partition of the parent nodes of $X_i$, $\pi(X_i)$, and let $\alpha_1^i, ..., \alpha_k^1$ denote the parameters of the additive decomposition. (The number of terms $k$ in the partition may vary across nodes.) We use Equations 2 and 3 to express the cross entropy given by Equation 5 as

$$I_{\text{Pr},\text{Pr}'} = \sum_{i=1}^n \sum_{x_i,\pi(X_i)} \Pr[x_i, \pi(X_i)] \left[ \log \Pr[x_i|\pi(X_i)] - \log \sum_{j=1}^k \alpha_j^i \Pr[x_i|S_j^i] \right]$$
$$= \sum_{i=1}^n I_i(\alpha_1^i, ..., \alpha_k^i),$$

where the the second sum in this expression is over all instantiations of node $X_i$ and its parents $\pi(X_i)$. Each term $I_i$ is a function of the parameters for node $X_i$ only. We compute the parameters which minimize the cross entropy by solving for $\alpha_j^i$ in

$$\frac{\partial I_i}{\partial \alpha_j^i} = 0, \qquad (6)$$

for all $i$ and $j$. Using the expression for $I_i$ in Equation 6 and since $\alpha_k = 1 - \sum_{j=1}^{k-1} \alpha_j$, we get

$$\sum_{x_i,\pi(X_i)} \Pr[x_i, \pi(X_i)] \frac{\Pr[x_i|S_j^i] - \Pr[x_i|S_k^i]}{\sum_{l=1}^k \alpha_l^i \Pr[x_i|S_l^i]} = 0. \quad (7)$$

Solution of Equation 7 is a difficult task when the number of parameters $l$ is large. We must search the $l$-dimensional unit cube for the solutions of this equation. Fortunately, $l$ is often small. Furthermore, to solve Equation 7 we must compute the probabilities $\Pr[x_i, \pi(X_i)]$ for all instantiations of $x_i$ and $\pi(X_i)$. In the worst case these probabilities cannot be computed exactly, and we must approximate them with a simulation algorithm. Nontheless, in many cases we can solve Equation 7 exactly to yield a set of parameters that minimizes the cross entropy between the full joint probabilities of BN and ABNM.

## 5   INFERENCE ALGORITHM

Both exact and approximate probabilistic inference in belief networks is NP-hard [6, 12], and therefore, intractable for sufficiently large belief networks. In this section, we develop an exact inference algorithm for ABNMs that exploits the additive decomposition. The run time of the algorithm depends on the decomposition of the ABNM, however, when we chose the decompositions thoughtfully, we render inference tractable in all cases.

The inference algorithm we present is similar to Cooper's nested dissection algorithm for probabilistic inference [5]. We decompose the belief network into subnetworks using the additive decomposition. We use the Lauritzen-Speiglehalter (L-S) algorithm [27] to perform inference on the subnetworks. The decomposition renders the subnetworks amenable to fast inference with the L-S algorithm. We then combine the results from each subnetwork inference to arrive at the desired inference probability.

We introduce the algorithm through an example. We assume familiarity with the L-S algorithm. Figure 2 gives a portion of ALARM [2], a belief network for patient monitoring in an intensive care unit; we call this subnetwork AlarmX. The L-S algorithm first builds a *triangulated graph* composed of *cliques*. Figure 3 shows the triangulated graph and shadows the nodes contained in the largest clique $C_0$: $\{x_0, x_3, x_5, x_6, x_8\}$. The algorithm constructs *clique marginals* for each clique. The clique marginals are probability distributions over all nodes of the clique—for example, the clique marginal for $C_0$ is the probability distribution $\Pr[x_0, x_3, x_5, x_6, x_8]$. For each clique, the L-S algorithm stores the table of clique marginals over all instantiations of the nodes in the clique. Thus, if $d_i$ denotes the number of values assumed by each node $X_i$ in the belief network, then for a clique $C$ comprised of nodes $x_1, ..., x_k$, the algorithm must store a table of clique marginal probabilities of size

$$N(C) = \prod_{i=1}^k d_i.$$

If each node in $C_0$ has five possible values, the the L-S algorithm stores a table of clique marginals for $C_0$ of size $5^5 = 3125$. The running time of the L-S algorithm is proportional to $N(C)$, evaluated at the largest clique $C$.

Nodes $x_1$, $x_2$, and $x_3$ are assumed to have the additive decompositions shown. The ABNM inference algorithm selects a decomposable node contained in the largest clique. The algorithm chooses $x_3$ in the example. The algorithm *dissects* the belief network at the chosen node to generate two belief networks $\text{BN}_\alpha$ and $\text{BN}_{1-\alpha}$. $\text{BN}_\alpha$ is obtained from BN by deleting the edges from nodes $x_5$ and $x_8$ to node $x_3$. $\text{BN}_{1-\alpha}$ is obtained from BN by deleting the edges from nodes $x_0$ and $x_6$ to node $x_3$. $\text{BN}_\alpha$ and $\text{BN}_{1-\alpha}$ and the corresponding triangulated graphs are shown in Figure 4. We observe that a single dissection reduces $N(C)$ from 3125 for BN, to 125 for both $\text{BN}_\alpha$ and $\text{BN}_{1-\alpha}$. If we want to compute the inference $\Pr[x_6|x_1]$, we use the L-S algorithm to compute the inferences $\Pr_\alpha[x_6|x_1]$ and

$\Pr_{1-\alpha}[x_6|x_1]$ in $BN_\alpha$ and $BN_{1-\alpha}$, respectively. We can verify readily that

$$\Pr[x_6|x_1] = \alpha \Pr_\alpha[x_6|x_1] + (1-\alpha)\Pr_{1-\alpha}[x_6|x_1].$$

We have reduced storage and computation from 3125 clique marginal probabilities to two tables of 125 probabilities.

We could continue to dissect both $BN_\alpha$ and $BN_{1-\alpha}$ at either $x_1$ or $x_2$. The process is identical, and if we were to dissect along all three nodes, we would generate eight sparse belief subnetworks. As a rule, however, we only dissect a node if it reduces the size of the largest clique. Thus, we avoid generating a large number of sparse belief subnetworks that we must store and evaluate each time we compute an inference.

## 6 IMPLEMENTATION RESULTS

We implemented our probabilistic inference algorithm for ABNMs. We present here results for AlarmX. To highlight the effects on the complexity of inference and on the cross entropy between the full joint probabilities, we assume that node $x_3$ is the only decomposable node.

To obtain the prior conditional probabilities $\Pr[x_3|x_0, x_6]$ and $\Pr[x_3|x_5, x_8]$ for AlarmX, we marginalized the conditional probability for $x_3$ in the full model. In general, we would assess these probabilities directly from the expert, or induce them from data by counting fractional occurences of the instantiations. The weight $\alpha_3$ denotes the contribution from $\Pr[x_3|x_0, x_6]$ to the conditional probability for $x_3$, and $1-\alpha_3$ denotes the contribution from $\Pr[x_3|x_5, x_8]$.

When $\alpha_3 = 0.485$, we obtain the minimum value for the cross entropy, $I_{\Pr,\Pr'} = 0.311079$. Recall that the cross entropy ranges from 0 to infinity, and it is identically zero if the two probability distributions are identical.

We compare the marginal probabilities we obtain with the full belief network and the ABNM for AlarmX. These probabilities were identical for all the nodes except node $X_1$. The full belief network gives marginal probabilities of 0.4444, 0.2723, and 0.2834, corresponding to a value of Low, Medium, and High for node $X_1$. For the same node, the ABNM gives marginal probabilities 0.4425, 0.2606, and 0.2969.

## 7 CONCLUSIONS

Like noisy-OR models and probabilistic similarity networks, ABNMs are approximate models that trade off predictive accuracy for speed and simplicity. Unlike the other methodologies, however, the ABNM methodology does not make as stringent an assumption about model structures or probabilities. When faced with a large, complex domain, a modeler can iteratively refine an ABNM by partitioning nodes that contribute significantly to intractability.

We have discussed the properties of ABNMs and have provided means for the estimation of their parameters. We have measured how well ABNMs model a domain by the cross entropy between the full joint probabilities of the ABNM and the full belief network. In an example ABNM, we found that the cross entropy was 0.311, and therefore, the ABNM provided a comparable model of the domain. Furthermore, the complexity of inference in the ABNM for our example was reduced by one order of magnitude. Future research objectives include (1) the development of search strategies for the partition that minimizes the cross entropy between the ABNM and the full belief network, (2) the extension of ABNMs to log-linear models [8], and (3) further tests and validation of the methodology.

## Acknowledgments

This work was supported by the National Science Foundation under grant IRI-9108385, by Rockwell International Science Center IR&D funds, and by the Stanford University CAMIS project under grant IP41LM05305 from the National Library of Medicine of the National Institutes of Health.